\documentclass[10pt,twocolumn,letterpaper]{article}

\usepackage{3dv}
\usepackage{times}
\usepackage{epsfig}
\usepackage{graphicx}
\usepackage{amsmath}
\usepackage{amssymb}
\usepackage{xspace}
\usepackage{hyperref}
\hypersetup{
    colorlinks=true,
    linkcolor=blue,
    filecolor=magenta,      
    urlcolor=cyan,
}

\usepackage{subcaption}
\usepackage{soul}
\usepackage{color, colortbl}
\usepackage{gensymb}
\usepackage{graphicx}

\makeatletter
\@namedef{ver@everyshi.sty}{}
\makeatother
\usepackage{tikz}
\usepackage{hhline, multirow}

\usepackage{booktabs}

\newcommand{\PreserveBackslash}[1]{\let\temp=\\#1\let\\=\temp}
\newcolumntype{C}[1]{>{\PreserveBackslash\centering}p{#1}}
\newcolumntype{R}[1]{>{\PreserveBackslash\raggedleft}p{#1}}
\newcolumntype{L}[1]{>{\PreserveBackslash\raggedright}p{#1}}

\definecolor{Red}{rgb}{1,0,0}
\definecolor{Gray}{rgb}{0.8,0.8,0.8}

\renewcommand{\deg}{\degree\xspace}

\newcommand{\MLP}[1]{MLP(#1)}
\newcommand{\MLPd}[1]{MLP$_{0.7}$(#1)}
\newcommand{\PointNet}[1]{PointNet(#1)}

\newcommand{\STAB}[1]{\begin{tabular}{@{}c@{}}#1\end{tabular}}
\newcommand{\abs}[1]{\left|#1\right|}

\newcommand{\PAR}[1]{\vskip2pt \noindent {\bf #1~}}

\newcommand{\method} {\mbox{AlignNet-3D}\xspace} 

\threedvfinalcopy %

\def\threedvPaperID{204} %

\ifthreedvfinal\fi
\begin{document}

\title{AlignNet-3D: Fast Point Cloud Registration of Partially Observed Objects} %

\author{
Johannes Gro{\ss}\\
{\tt\small johannes.gross1@rwth-aachen.de}
\and
Aljo\u{s}a O\u{s}ep\\
{\tt\small osep@vision.rwth-aachen.de}\\\\
Computer Vision Group\\
RWTH Aachen University
\and
Bastian Leibe\\
{\tt\small leibe@vision.rwth-aachen.de}
}

\maketitle
\thispagestyle{empty}

\begin{abstract}
  Methods tackling multi-object tracking need to estimate the number of targets in the sensing area as well as to estimate their continuous state.
  While the majority of existing methods focus on data association, precise state (3D pose) estimation is often only coarsely estimated by approximating targets with centroids or (3D) bounding boxes.
  However, in automotive scenarios, motion perception of surrounding agents is critical and inaccuracies in the vehicle close-range can have catastrophic consequences.
  In this work, we focus on precise 3D track state estimation and propose a learning-based approach for object-centric relative motion estimation of partially observed objects.
  Instead of approximating targets with their centroids, our approach is capable of utilizing noisy 3D point segments of objects to estimate their motion.
  To that end, we propose a simple, yet effective and efficient network, \method, that learns to align point clouds. %
  Our evaluation on two different datasets demonstrates that our method outperforms computationally expensive, global 3D registration methods while being significantly more efficient. %
  We make our data, code, and models available at~\url{https://www.vision.rwth-aachen.de/page/alignnet}. 
\end{abstract}

\section{Introduction}
Multi-object tracking (MOT) is a well-established field with a long research history. Originating in point-based object tracking in aircraft and naval scenarios based on RADAR sensors~\cite{Reid79TAC}, it nowadays plays an essential role in mobile perception for autonomous vehicles.
Through tracking, intelligent vehicles can become aware of surrounding object instances and estimate their current pose and extent in 3D space and time, allowing them to predict future motion and to react in time to potentially harmful situations.

\begin{figure}[t]
    \centering
      \includegraphics[width=0.9\linewidth]{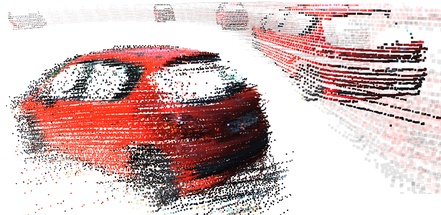}
      \vspace{-0.95pt}
      \caption{Given LiDAR segments of an object, captured at different time-steps (some frames are highlighted for illustrative purposes), our method learns to align these segments to estimate the relative motion. Shown above are all segments transformed to a common coordinate frame using our predicted frame-to-frame alignments.}
      \label{subfig:intro1}
    \vspace{-2pt}
    \label{fig:intro}
\end{figure}

In the context of autonomous driving, the vast majority of existing vision and LiDAR based methods focus on the detection of targets in the scene and on the subsequent data association~\cite{ButtCollins13CVPR, Choi15ICCV, Zhang08CVPR, Yoon15WACV, Andriyenko12CVPR, Milan14TPAMI}. Recent learning-based trends aim at learning data association affinity functions~\cite{Xiang15ICCV, Schulter17CVPR}, at obtaining temporally-stable detections~\cite{Feichtenhofer17ICCV, Voigtlaender19CVPR, Kang17CVPR} and at tackling joint segmentation and tracking~\cite{Voigtlaender19CVPR}.
Even though high-level path planning requires motion perception in 3D space, the task of continuous 3D state estimation of tracked targets is often neglected in existing approaches.
Methods based on network flow~\cite{Zhang08CVPR, Schulter17CVPR, Lenz15ICCV, Pirsiavash11CVPR} only focus on discrete optimization and do not estimate the continuous state of targets.
Several approaches cast state estimation as inference in linear dynamical systems such as Kalman filters. Most often, the target state is parametrized as a 2D bounding box in the image domain (vision-based methods~\cite{Kim15ICCV, Lenz15ICCV, Milan14TPAMI, Sadeghian17ICCV}) or approximated by the center of mass or 3D bounding box in case of stereo-based~\cite{Leibe08TPAMI, Osep17ICRA, Mitzel10ECCV, Ess09PAMI} or LiDAR-based methods~\cite{Teichman11ICRA, Petrovskaya09AR, Dewan15ICRA, Kaestner12ICRA, Moosmann13ICRA}.
While such coarse approximations are perfectly reasonable in case the target is not in our direct proximity, it is imperative that we estimate motion precisely in the vehicle close-range -- imprecisions in this range can lead to catastrophic consequences.

An exception is the work of~\cite{Held14RSS}, which proposes to estimate object velocity by sampling 3D positions on the ground plane that yield small point-to-point distance errors between consecutive LiDAR scans of objects and thus takes full 3D shape measurements into account for trajectory estimation.
However, this approach can fail in case it does not have a good initial pose estimate.

\begin{figure}[t]
  \begin{center}
      \includegraphics[width=1.0\linewidth, clip, trim=0 11cm 2cm 0]{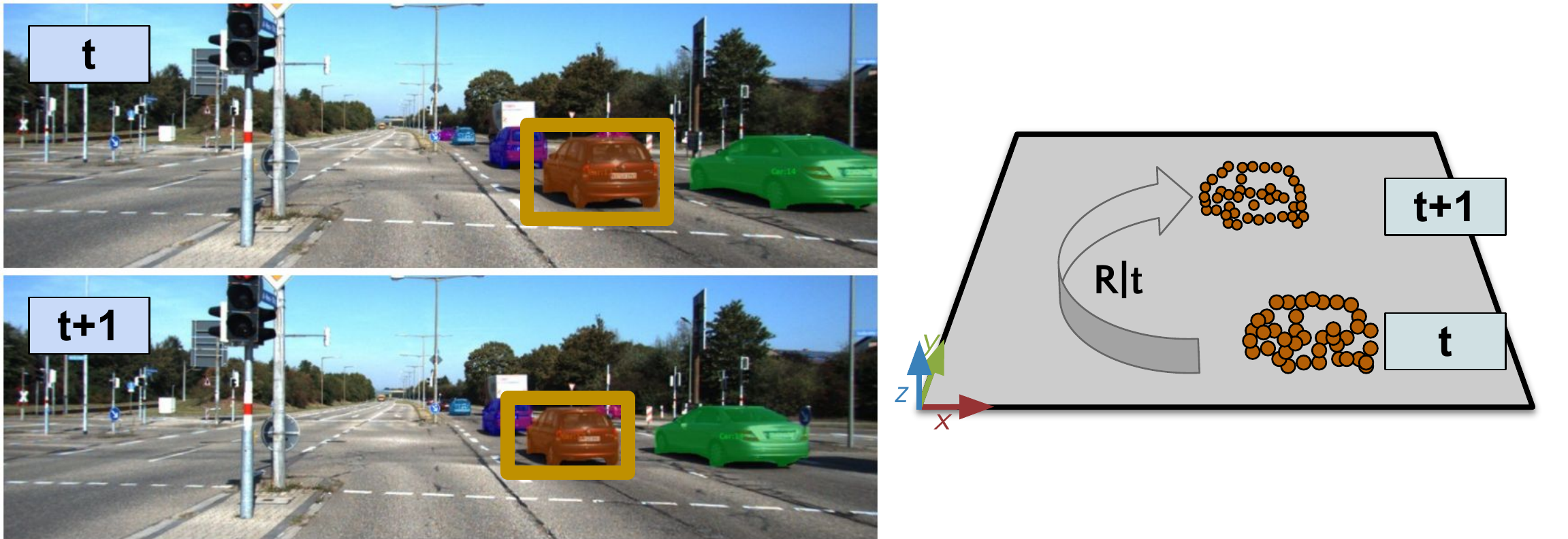}
  \end{center}
  \vspace{-15pt}
  \caption{High-level overview of our approach. Given two point clouds of an object at two different time-steps, we precisely estimate the relative motion of an object by aligning their observed 3D point clouds.}
\label{fig:hilevel-overview}
\end{figure}

In this work, we explicitly focus on the state estimation task and assume that association between the point segments (\ie, point segment identity) is given.
Instead of performing direct optimization~\cite{Held14RSS, Ushani15IROS, zhou2016fast, Besl92PAMI}, we propose a learning-based approach, that optimizes weights of a neural network, such that, given two LiDAR point clouds, the error in the estimated relative motion is minimized. Such an approach has the advantage that it can improve its performance with an increasing amount of training data. %
As in~\cite{Held14RSS, Ushani15IROS}, our approach goes beyond simple centroid or bounding-box based approximation and is capable of utilizing full (noisy and possibly imprecise) 3D point segments of observed targets to precisely estimate the motion of objects.
Our main contribution is a simple, yet effective and efficient 3D scan alignment network -- \method{} -- that learns to align point clouds in a data-driven fashion to estimate precise relative transformations between 3D point sets.

More precisely, given two LiDAR point segments that measure object shape at different times, our network estimates relative motion between scans (see Fig.~\ref{fig:hilevel-overview}) by learning to align the observed point sets.
Contrary to purely geometric methods, our method is robust to large temporal gaps, as we experimentally demonstrate. 
This makes our method applicable not only to object tracking but potentially also to tasks such as 3D reconstruction and shape completion.

To study the alignment network problem in a well-controlled setting, we generate a synthetic dataset for training and evaluation of our method by sampling points from CAD models so that they mimic characteristics of the LiDAR sensor.
To show that our method generalizes well to real-world scenarios, we then evaluate our method on the KITTI tracking~\cite{Geiger12CVPR} dataset.
Our method outperforms several variants of the ICP algorithm~\cite{Besl92PAMI} and performs as well as the computationally expensive global point cloud registration method~\cite{zhou2016fast}, while being significantly faster.
In addition, we obtain lower velocity estimation error compared to a strong tracking-based baseline~\cite{Held14RSS}.

In summary, we make the following contributions: 1)~we propose an efficient, learning-based method for registration of partially-observed 3D object shapes, captured with 3D sensors.
2) to study this problem in a well-controlled setting, we create a new synthetic dataset based on ModelNet40~\cite{Wu15CVPR} by sampling points from CAD models so that they mimic characteristics of the LiDAR sensor and new evaluation metrics.
We compare our method with global registration methods and approaches for LiDAR-based state estimation and demonstrate excellent performance on synthetic, as well as real-world datasets.
3) we make our code, experiments and synthetic dataset publicly available. %

\section{Related Work}

Methods for multi-object tracking need to identify and locate an unknown number of objects and their trajectories over time and to accurately estimate their current and future states (\eg, position, velocity, orientation).
Existing LiDAR~\cite{Teichman11ICRA, Petrovskaya09AR, Dewan15ICRA, Kaestner12ICRA, Moosmann13ICRA, Frossard18ICRA} and vision based methods~\cite{Kim15ICCV, Lenz15ICCV, Milan14TPAMI, Sadeghian17ICCV} mainly focus on the task of estimating the number of targets and on the temporal association of measurements~\cite{Zhang08CVPR, Schulter17CVPR, Lenz15ICCV, Pirsiavash11CVPR}.

\PAR{State Estimation.}~State estimation is often posed as inference on a Markov chain using, \eg, Kalman filters~\cite{Leibe08TPAMI, Osep17ICRA, Mitzel10ECCV, Ess09PAMI} or particle filters~\cite{Okuma04ECCV, Danescu12ITSM, Fu18ACCESS}, with recent trends towards data-driven sequence modeling using recurrent neural networks, such as Long-Short Term Memory cells (LSTMs)~\cite{Sadeghian17ICCV}.
The majority of these methods for 3D object tracking using LiDAR and RGB-D sensors approximate the object state for trajectory estimation with centroids or bounding boxes~\cite{Teichman11ICRA, Petrovskaya09AR, Dewan15ICRA, Moosmann13ICRA, Osep17ICRA, Mitzel10ECCV}.
This can lead to imprecise trajectory and state estimation, especially in close proximity of the vehicle, when objects are occluded and measurements are truncated.

Exceptions are~\cite{Held14RSS, Mitzel12ECCV, Ushani15IROS, Osep16ICRA}, which utilize full 3D point clouds, representing surface measurements of objects in order to estimate the motion of targets precisely.
Held \etal~\cite{Held14RSS} sample positions on a ground-plane estimate in a coarse-to-fine scheme to obtain an estimate that minimizes a point-to-point based distance error between point sets. Mitzel~\etal~\cite{Mitzel12ECCV} and O\u{s}ep~\etal~\cite{Osep16ICRA} perform frame-to-frame ICP-based alignment between a fixed-dimensional representation of the tracked target and stereo measurements of the object surface.
Ushani \etal~\cite{Ushani15IROS} simultaneously optimize the object trajectory and 3D object shape over time using iterative batch optimization.
All aforementioned methods optimize transformations directly, whereas our method \emph{learns} to align objects. Thus, it can benefit from additional training data and learn to be robust against occlusions and truncations.

\PAR{Object Registration.}~The task of relative motion estimation can be cast in terms of object registration.
The most commonly used algorithm for point set registration is the iterative closest point (ICP) algorithm~\cite{Besl92PAMI}.
The main idea behind ICP is to minimize a sum-of-squared-distances (SSD) error between the model and target point sets.
This optimization is performed iteratively by 1) finding closest point pairs between the model and target and 2) estimating a transformation that minimizes the overall distance between these pairs.
Existing methods mainly differ by different strategies on point matching (\eg, closest point (point-to-point), or using normal information (point-to-plane)) and by their choice of error metric and optimization strategy (an SSD based error leads to closed-form solutions~\cite{Besl92PAMI}). For an exhaustive overview of ICP variants, we refer to~\cite{Rusinkiewicz013DIM}.

The ICP algorithm needs a good initial pose estimate in order to converge to a correct solution.
To alleviate this problem, Rusu~\etal~\cite{Rusu09ICRA} propose Global ICP.
Here, first a coarse alignment is estimated using RANSAC by matching FPFH~\cite{Rusu09ICRA} features. 
However, this approach is too computationally demanding to be applicable for real-time tracking.
Go-ICP~\cite{Yang15TPAMI} is a global registration method that overcomes the need for initialization with a global branch-and-bound search in SE(3), which guarantees the convergence to a global optimum.
On our data, the computation time of the (single-threaded) implementation is up to 20 seconds per transform, which is not suitable for a real-time multi-object tracker.
The Fast Global Registration (FGR) approach by~\cite{zhou2016fast} provides good initializations for a local ICP refinement, but it is still too slow to be applicable to real-time multi-object tracking scenarios.
We evaluate our method against both local and global ICP as well as FGR.
Recent methods use learned features and feature matching algorithms~\cite{Zeng17CVPR, Yew18ECCV, Deng18CVPR}, still paired with iterative RANSAC schemes.

\PAR{Parallel Work.}~Most similar to our method is the recently proposed data-driven~\cite{Goforth19CVPR} point cloud registration algorithm. However, their approach is iterative, based on a recurrent deep neural network.
In contrast, we focus on efficient registration for tracking scenarios. Instead of an iterative registration scheme, we propose a single-stage approach, that first learns to coarsely align point sets by estimating their canonical orientation, followed by a refined transformation estimation.
Recent work of~\cite{Giancola18CVPR} leverages shape completion for the task of 3D LiDAR-sequence segmentation.
Different to our method, their goal is to localize the target (assuming segmentation given in the first frame) in the following LiDAR scans.
This is similar to the task of video-object segmentation~\cite{Caelles18arXiv}.

\section{Method}

We formulate track state estimation as a task of estimating a relative transform between two 3D point sets.
These two point sets correspond to a (sparse) measurement of the surface of an object, captured at different time steps (see Fig.~\ref{fig:hilevel-overview}).
In this work, we assume we are given a segmentation of the object, as well as the association between the point segments, \ie, we assume we have a high-level tracker that provides unique identity assignments to 3D point sets.
In this context, several existing methods can be used, both LiDAR~\cite{Frossard18ICRA, Teichman11ICRA} or RGB based~\cite{Voigtlaender19CVPR}.
As we mainly target (automotive) real-time tracking scenarios, we constrain the pose transform prediction to translation on the ground plane (equivalent to the $xy$-plane in all experiments) and relative rotation around the $z$-axis.
We train our network (Sec.~\ref{sec:arch}) using 3D point cloud pairs that represent object surface measurements at different time steps, together with a ground truth alignment in the form of a 2D ground plane translation and a relative angle.
Such training triplets can be generated synthetically (using CAD models, \eg~\cite{Wu15CVPR}) or from the real-world KITTI tracking dataset~\cite{Geiger12CVPR}, where 3D bounding boxes of objects and object identity information are provided.
We discuss datasets and training data generation in Sec.~\ref{sec:datasets}.

\begin{figure}[t]
    \centering
    \begin{subfigure}[b]{1.0\linewidth}
        \includegraphics[width=1.0\linewidth, clip, trim=0 11.5cm 1cm 0]{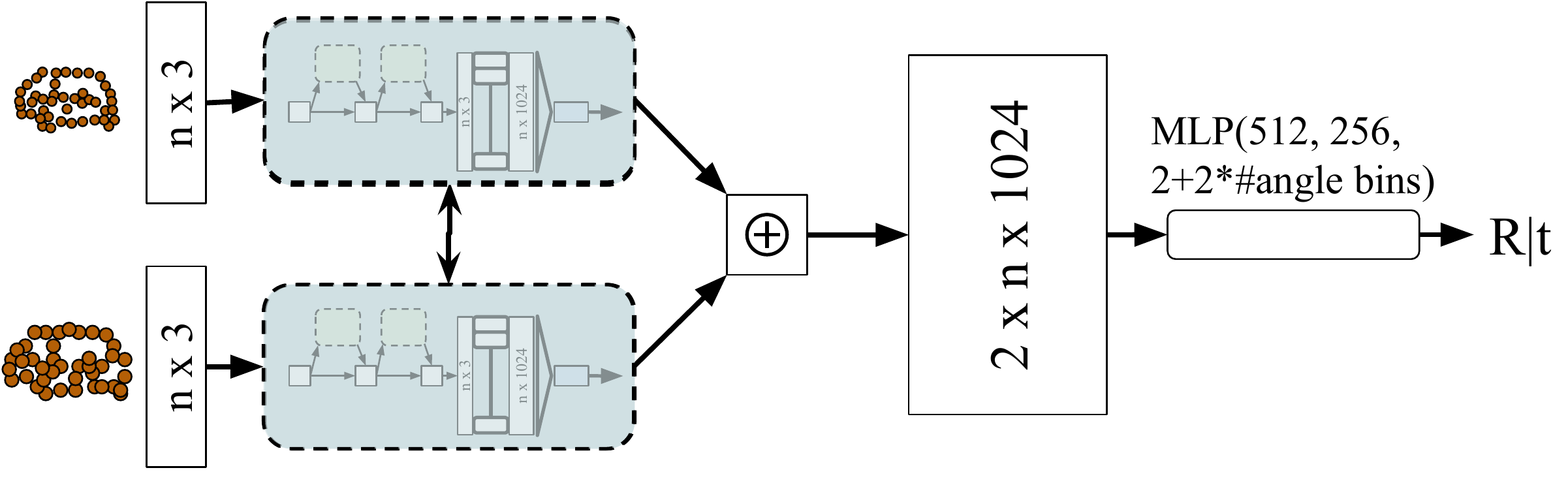}%
        \vspace{-4pt}
        \caption{A high-level overview of our network. We transform two given point segments using their (estimated) canonical pose and extract their embeddings. %
        Then, we concatenate both embeddings and use an MLP to obtain a refined estimate of the alignment.}
        \label{subfig:hi-level}
    \end{subfigure}
    \begin{subfigure}[b]{1.0\linewidth}
        \includegraphics[width=1.0\linewidth, clip, trim=0 13cm 1cm 0]{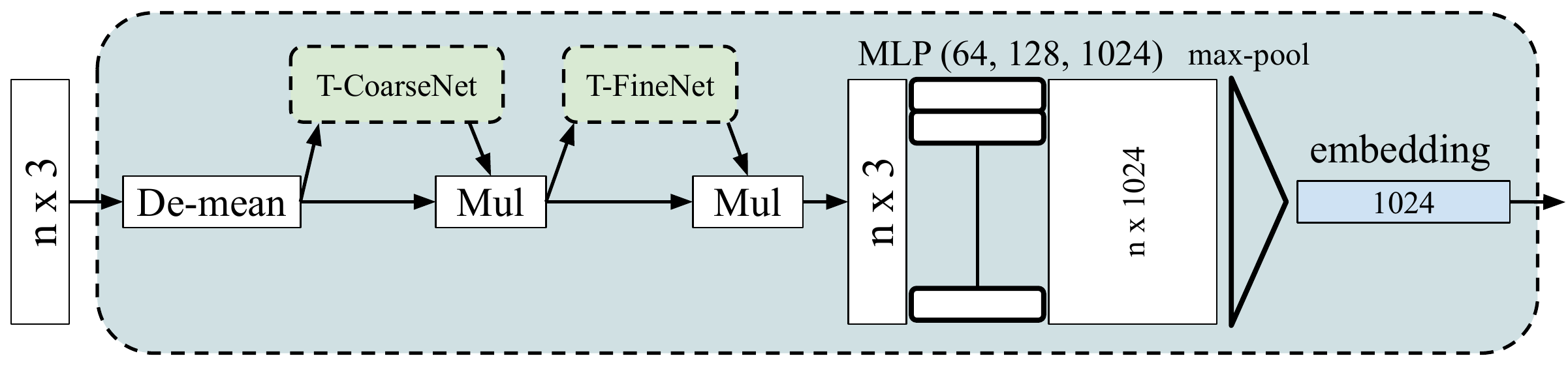}%
        \vspace{-4pt}
        \caption{The CanonicalNet, a branch of the siamese network, estimates a coarse amodal center of the object, followed by an estimate of a refined object center and its orientation.
        Finally, we extract a PointNet embedding of the normalized point cloud.}
        \label{subfig:canonicalnet}
    \end{subfigure}
    \vspace*{-6mm}
    \caption{Our network architecture:~(\subref{subfig:hi-level}) high-level overview of the proposed siamese network and~(\subref{subfig:canonicalnet}) sub-network for canonical pose estimation.}
    \label{fig:network}
\end{figure}

\subsection{Network Architecture}
\label{sec:arch}

Input to our network (Fig.~\ref{subfig:hi-level}) are two 3D point clouds that represent surface measurements of an object, captured at two different time steps.
To ensure both point clouds are of the same size, we randomly sample $n$ points from each point cloud.
Both point clouds are input to CanonicalNet (Fig.~\ref{subfig:canonicalnet}), two branches of a siamese network, which transform the point clouds to a canonical pose and compute a fixed-dimensional feature vector for each point cloud.
To obtain a refined alignment estimate, we concatenate embeddings of both point clouds and use a multi-layer perceptron~(MLP) to produce the final transform.

To process the sparse point cloud data (within a branch of the siamese network), we utilize the PointNet architecture of \cite{Qi2017CVPR}.
In the following we denote a PointNet that consists of a per-point MLP of layer sizes 64, 128 and 256, followed by max-pooling, as \PointNet{64, 128, 256}. Multi-layer perceptrons with 512 and 256 hidden layers are abbreviated as \MLP{512, 256}, or \MLPd{512, 256} if a dropout of 70\% is applied after the last layer. All hidden layers use the ReLU activation function.

\PAR{CanonicalNet.}~We visualize our CanonicalNet architecture in Fig.~\ref{subfig:canonicalnet}.
We first normalize the input point cloud by moving its centroid to the origin.
Using T-CoarseNet (\PointNet{64, 128, 256}, followed by \MLPd{512, 256}) we predict an amodal object center and bring the object closer to a canonical pose by moving the predicted center to the origin.
After the coarse center estimate, we use T-FineNet (\PointNet{64, 128, 512}, followed by \MLPd{512, 256}) to refine the object center and additionally predict a canonical orientation.
To estimate orientation we use a hybrid approach (similar to~\cite{Qi2018CVPR}) for angle prediction.
Instead of directly regressing the angle to the canonical orientation, we predict a classification into one of 50 equidistant angle bins between 0 and $2\pi$.
Additionally, we predict an angle residual for every angle bin, so that the final angle prediction is computed as $\alpha=i \cdot \frac{2\pi}{\text{\#bins}} + \text{res}_i$, where $i$ is the predicted angle bin and $\text{res}_i$ is the respective predicted residual. The output layer size of an MLP predicting a translation and an angle is thus $2+2\cdot \text{\#bins}$.
We re-normalize the point cloud by moving the new amodal center to the origin, followed by a rotation by $-\alpha$.
The final point cloud embedding is predicted by a \PointNet{64, 128, 1024} network.

\PAR{Final Alignment.}~Finally, we concatenate the point cloud embeddings computed by the siamese branches (see Fig.~\ref{subfig:hi-level}).
We input the combined feature vector of size $2048$ to a final \MLPd{512, 256}, which predicts refined translation and angle to precisely align the two point clouds in their canonical pose estimates.
During inference, with the canonical transforms $T_1$ and $T_2$ computed by the siamese branches and the final alignment $T_f$, we compute the overall alignment of the two input point clouds as $T_1T_fT_2^{-1}$.

\subsection{Loss Function}

All stages of our pipeline are fully supervised. In \emph{stage~1}, we predict an amodal center with our first transformer network, T-CoarseNet (Fig.~\ref{subfig:canonicalnet}).
As target center we use the center of the 3D bounding box in our datasets based on KITTI, and the mesh origin in our synthetic datasets based on ModelNet~\cite{Wu15CVPR}.
In \emph{stage~2}, our second transformer network, T-FineNet, predicts a refined amodal center and the deviation from a canonical orientation.
The target rotation angle is the mesh or annotated 3D bounding box orientation.
In \emph{stage 3}, we predict the remaining translation and rotation needed to align the point clouds from the concatenated embeddings (see Fig.~\ref{subfig:hi-level}). Here we penalize the deviation from the ground truth remaining transform as follows. %

Translation and orientation deviations are penalized by our loss function independently, which is heavily inspired by \cite{Qi2018CVPR}. For the translation deviation, the Huber loss function~\cite{Huber64} with $\delta=1$ is used, except for the last stage, where $\delta=2$, as in \cite{Qi2018CVPR}. The Huber loss function $L_\delta(x)$ is defined as $\frac12x^2$ for $\abs{x}\leq\delta$ and $\delta\abs{x}-\frac12\delta^2$ otherwise. The translation loss is abbreviated as  $L_\text{transl}$. The angle loss $L_\text{angle}$ comprises of the cross entropy loss $L_\text{cls}$ for the angle bin classification, and a Huber loss $L_\text{reg}$ for the residual corresponding to the ground truth angle bin. Residuals are predicted normalized within $[-1, 1]$, corresponding to angles in $[-\beta/2, \beta/2]$ with the angle bin size $\beta=\frac{2\pi}{\text{\#bins}}$. The predicted normalized residuals are penalized by the Huber loss function with normalized target residual angles and $\delta=1$. The classification and regression losses are combined to
\setlength{\abovedisplayskip}{5pt}
\setlength{\belowdisplayskip}{5pt}
\setlength{\abovedisplayshortskip}{5pt}
\setlength{\belowdisplayshortskip}{5pt}
\begin{equation}
L_\text{angle}=L_\text{cls}+20\cdot L_\text{reg}.
\end{equation}
The overall loss is computed as:
\begin{equation}
L=L_\text{transl, overall}+\lambda_2\cdot L_\text{angle, overall},
\end{equation}
with
\begin{align}
&L_\text{transl, overall} = \lambda_1(L_\text{transl, s1}+L_\text{transl, s2})+L_\text{transl, s3} \label{eq:1},\\
&L_\text{angle, overall} = \lambda_1 L_\text{angle, s2}+L_\text{angle, s3}\label{eq:2},
\end{align}
where the individual losses of stage 1 and 2 are averaged between the siamese branches. For the datasets based on KITTI, the hyperparameters $\lambda_1, \lambda_2$ are set to $0.5$ each. $\lambda_1$~weights the losses of the earlier stages, $\lambda_2$~balances the translation and angle losses.
For the synthetic datasets, $\lambda_2$ is set to $1.0$.

\subsection{Training}
We train our network with the Adam optimizer for 200 epochs, with a learning rate of $0.005$, which decays by 0.5 every 30 epochs. We use an input point cloud size of $n=512$ (randomly sampled from the original point clouds), with a batch size of 128. We do not use any color channels.
During training, point clouds are augmented by adding noise to every point and coordinate, sampled from $\mathcal N(0, \sigma)$ with $\sigma=0.01$, but clipped to $\left[-0.05, 0.05\right]$. We use batch normalization for all layers, with batch norm decay starting at 0.5, decaying by 0.5 every 30 epochs (although clipped at 0.99). We have tuned all hyperparameters on the validation sets, while reporting our numbers on the test sets.

\begin{table*}[t]
  \centering\footnotesize\begin{tabular}{l|rrrrr|rrrrr|r}
  \multicolumn{1}{c}{} & \multicolumn{5}{c}{distance $<$ 80m} & \multicolumn{5}{c}{distance $<$ 20m} \\
\hline
                &    2cm,1$^\circ$ &   10cm,5$^\circ$ &   20cm,10$^\circ$ &         RMSE $t$ &            RMSE $R$ &    2cm,1$^\circ$ &   10cm,5$^\circ$ &   20cm,10$^\circ$ &         RMSE $t$ &            RMSE $R$ & time/transform      \\
\hline
 Global ICP     &           4.00\% &          25.90\% &           35.30\% &          0.91m &         41.19$^\circ$ &          11.26\% &          52.81\% &           57.14\% &          0.58m &         38.74$^\circ$ &        2315.57ms \\
 Global ICP+p2p &          18.20\% &          32.30\% &           39.10\% &          0.90m &         40.74$^\circ$ &          42.42\% &          59.31\% &           61.04\% &          0.58m &         37.35$^\circ$ &        2341.20ms \\
\hline
 ICP p2p        &           8.10\% &          14.30\% &           17.90\% &          0.41m &         49.53$^\circ$ &          13.42\% &          18.61\% &           22.08\% &          0.48m &         49.22$^\circ$ &          27.88ms \\
 FGR            &           5.10\% &          12.30\% &           18.40\% &          0.44m &         46.52$^\circ$ &          18.61\% &          35.50\% &           41.56\% &          0.55m &         36.43$^\circ$ &          13.67ms \\
 FGR+p2p        &          14.40\% &          21.10\% &           25.00\% &          0.48m &         46.28$^\circ$ & \textbf{40.26\%} &          47.19\% &           50.22\% &          0.56m &         36.14$^\circ$ &          34.53ms \\
 Ours           &           0.60\% &          34.90\% &  \textbf{74.90\%} & \textbf{0.19m} & \textbf{5.16$^\circ$} &           0.43\% &          38.53\% &  \textbf{77.92\%} & \textbf{0.20m} & \textbf{4.73$^\circ$} &  \textbf{1.22ms} \\
 Ours+p2p       & \textbf{17.60\%} & \textbf{48.50\%} &           73.00\% &          0.21m &          6.62$^\circ$ &          25.54\% & \textbf{49.78\%} &           73.16\% &          0.24m &          8.16$^\circ$ &          22.97ms \\
\hline
\end{tabular}

  \vspace{-6pt}
  \caption{Registration results on \texttt{SynthCars}, a dataset of simulated LiDAR scans of car meshes in random poses.}
  \label{tab:SynthCars}
\end{table*}
\begin{table*}[t]
  \centering\footnotesize\begin{tabular}{l|rrrrr|rrrrr|r}
  \multicolumn{1}{c}{} & \multicolumn{5}{c}{distance $<$ 80m} & \multicolumn{5}{c}{distance $<$ 20m} \\
  \hline
                &   2cm,1$^\circ$ &   10cm,5$^\circ$ &   20cm,10$^\circ$ &         RMSE $t$ &                 RMSE $R$                 & 2cm,1$^\circ$    & 10cm,5$^\circ$   & 20cm,10$^\circ$   & RMSE $t$ &                 RMSE $R$                 & time/transform       \\
\hline
 Global ICP     &           3.40\% &          24.80\% &           33.50\% &          0.80m &          40.68$^\circ$ &          10.31\% &          53.36\% &           60.09\% &          0.56m &          35.35$^\circ$ &        2169.69ms \\
 Global ICP+p2p &          14.50\% &          32.30\% &           39.60\% &          0.79m &          39.93$^\circ$ &          38.12\% &          63.68\% &           66.82\% &          0.53m &          33.25$^\circ$ &        2188.65ms \\
\hline
 ICP p2p        &           7.50\% &          16.50\% &           22.40\% &          0.38m &          49.72$^\circ$ &          15.25\% &          26.01\% &           31.39\% &          0.48m &          48.11$^\circ$ &          26.56ms \\
 FGR            &           4.10\% &          11.60\% &           17.50\% &          0.42m &          48.44$^\circ$ &          16.59\% &          34.53\% &           42.15\% &          0.49m &          38.56$^\circ$ &          14.83ms \\
 FGR+p2p        & \textbf{12.00\%} &          20.80\% &           26.60\% &          0.45m &          46.52$^\circ$ & \textbf{35.87\%} & \textbf{47.98\%} &           52.91\% &          0.51m &          36.41$^\circ$ &          35.14ms \\
 Ours           &           0.30\% &          16.00\% &           45.80\% & \textbf{0.29m} &          19.41$^\circ$ &           0.45\% &          17.94\% &           50.22\% & \textbf{0.33m} &          16.44$^\circ$ &  \textbf{1.22ms} \\
 Ours+p2p       &           6.20\% & \textbf{30.10\%} &  \textbf{53.10\%} &          0.31m & \textbf{18.70$^\circ$} &          10.76\% &          36.77\% &  \textbf{61.43\%} &          0.34m & \textbf{15.09$^\circ$} &          12.32ms \\
\hline
\end{tabular}

  \vspace{-6pt}
  \caption{Registration results on \texttt{SynthCarsPersons}, containing about 80\%/20\% simulated scans of car/person meshes.}
  \label{tab:SynthCarsPersons}
\end{table*}
\begin{table*}[t]
  \footnotesize\begin{tabular}{cl|rrrrr|rrrrr|r}
  \multicolumn{2}{c}{} & \multicolumn{5}{c}{distance $<$ 80m} & \multicolumn{5}{c}{distance $<$ 20m} \\
  \cline{2-13}
            &&   2cm,1$^\circ$ &   10cm,5$^\circ$ &   20cm,10$^\circ$ &         RMSE $t$ &                 RMSE $R$                 & 2cm,1$^\circ$    & 10cm,5$^\circ$   & 20cm,10$^\circ$   & RMSE $t$ &                 RMSE $R$                 & time/transform       \\
\cline{2-13}
\multirow{7}{*}{\STAB{\rotatebox[origin=c]{90}{}}} & 
     Global ICP &          3.10\% &          16.70\% &           23.00\% &          0.70m &          67.76$^\circ$ &           8.65\% &          30.83\% &           34.96\% &          0.39m &          55.16$^\circ$ & 2141.82ms        \\
    & Global ICP+p2p &         10.70\% &          22.50\% &           28.20\% &          0.68m &          68.61$^\circ$ &          24.44\% &          38.35\% &           40.98\% &          0.36m &          55.06$^\circ$ & 2172.83ms        \\
\cline{2-13}
    &        ICP p2p &          7.20\% &          16.70\% &           22.20\% & \textbf{0.35m} & \textbf{51.81$^\circ$} &          15.41\% &          24.44\% &           28.95\% & \textbf{0.37m} & \textbf{50.60$^\circ$} & 21.31ms          \\
    &            FGR &          1.90\% &           7.50\% &           13.00\% &          0.47m &          61.01$^\circ$ &           6.39\% &          15.41\% &           21.80\% &          0.40m &          62.98$^\circ$ & 18.44ms          \\
    &        FGR+p2p &          8.70\% &          18.30\% &           23.80\% &          0.50m &          61.41$^\circ$ & \textbf{21.43\%} & \textbf{31.95\%} &  \textbf{37.22\%} &          0.45m &          61.95$^\circ$ & 37.84ms          \\
    &           Ours &          0.50\% &          11.30\% &           24.60\% &          0.41m &          93.81$^\circ$ &           0.75\% &          15.04\% &           27.07\% & \textbf{0.37m} &          91.64$^\circ$ & \textbf{1.22ms}  \\
    &       Ours+p2p & \textbf{9.50\%} & \textbf{23.50\%} &  \textbf{30.40\%} &          0.47m &          92.20$^\circ$ &          16.92\% &          30.83\% &           35.71\% &          0.42m &          90.99$^\circ$ & 23.35ms          \\
\cline{2-13}
\end{tabular}

  \vspace{-6pt}
  \caption{Registration results on \texttt{Synth20}, containing simulated scans of meshes of 20 different classes of ModelNet40~\cite{Wu15CVPR}. Here the angle deviation to the actual object headings is considered, instead of their heading axes.}
  \label{tab:Synth20}
\end{table*}
\begin{table*}[t]
  \footnotesize\begin{tabular}{cl|rrrrr|rrrrr|r}
  \multicolumn{2}{c}{} & \multicolumn{5}{c}{distance $<$ 80m} & \multicolumn{5}{c}{distance $<$ 20m} \\
  \cline{2-13}
            &&   2cm,1$^\circ$ &   10cm,5$^\circ$ &   20cm,10$^\circ$ &         RMSE $t$ &                 RMSE $R$                 & 2cm,1$^\circ$    & 10cm,5$^\circ$   & 20cm,10$^\circ$   & RMSE $t$ &                 RMSE $R$                 & time/transform       \\
\cline{2-13}
\multirow{7}{*}{\STAB{\rotatebox[origin=c]{90}{}}} & 
     Global ICP &           2.70\% &          15.90\% &           22.60\% &          0.81m &          70.02$^\circ$ &           8.07\% &          30.04\% &           34.98\% &          0.53m &          53.24$^\circ$ & 2185.02ms        \\
    & Global ICP+p2p &          12.60\% &          23.90\% &           27.90\% &          0.82m &          70.59$^\circ$ &          31.39\% &          39.46\% &           40.36\% &          0.55m &          52.54$^\circ$ & 2207.81ms        \\
\cline{2-13}
    &        ICP p2p &           7.90\% &          19.30\% &           23.20\% & \textbf{0.47m} & \textbf{52.62$^\circ$} &          13.45\% &          24.66\% &           27.80\% & \textbf{0.52m} & \textbf{50.80$^\circ$} & 27.64ms          \\
    &            FGR &           1.30\% &           8.10\% &           14.10\% &          0.61m &          66.73$^\circ$ &           4.48\% &          16.59\% &           26.91\% &          0.65m &          74.55$^\circ$ & 22.77ms          \\
    &        FGR+p2p &           9.60\% &          21.30\% &           25.00\% &          0.64m &          68.18$^\circ$ &          22.87\% &          36.32\% &           40.81\% &          0.65m &          75.68$^\circ$ & 48.60ms          \\
    &           Ours &           0.30\% &          10.20\% &           25.20\% &          1.00m &          91.50$^\circ$ &           0.00\% &          10.31\% &           27.80\% &          0.53m &          88.21$^\circ$ & \textbf{1.22ms}  \\
    &       Ours+p2p & \textbf{12.20\%} & \textbf{28.30\%} &  \textbf{35.00\%} &          1.04m &          91.98$^\circ$ & \textbf{23.32\%} & \textbf{37.67\%} &  \textbf{43.05\%} &          0.65m &          88.84$^\circ$ & 27.12ms          \\
\cline{2-13}
\end{tabular}

  \vspace{-6pt}
  \caption{Registration results on \texttt{Synth20others}, with simulated scans of the remaining 20 classes of ModelNet40~\cite{Wu15CVPR}. Again, the deviation to the actual object headings is used.}
  \label{tab:Synth20others}
\end{table*}
\begin{table*}[t]
  \centering\setlength{\tabcolsep}{3pt}\footnotesize\begin{tabular}{l|rrrrr|rrrrr|r}
  \multicolumn{1}{c}{} & \multicolumn{5}{c}{distance $<$ 80m} & \multicolumn{5}{c}{distance $<$ 20m} \\
\hline
                &    2cm,1$^\circ$ &   10cm,5$^\circ$ &   20cm,10$^\circ$ &         RMSE $t$ &            RMSE $R$ &    2cm,1$^\circ$ &   10cm,5$^\circ$ &   20cm,10$^\circ$ &         RMSE $t$ &            RMSE $R$ & time/transform      \\
\hline
 Global ICP            &          16.27\% &          67.89\% &           77.47\% &          0.67m &          6.65$^\circ$ &          17.46\% &          84.62\% &           91.72\% &          0.47m &          5.47$^\circ$ &        1766.97ms \\
 Global ICP+p2p        &          27.47\% &          71.58\% &           78.65\% &          0.66m &          6.60$^\circ$ &          36.98\% &          89.64\% &           93.49\% &          0.45m &          4.85$^\circ$ &        1773.58ms \\
\hline
 ICP p2p               & \textbf{25.85\%} & \textbf{67.67\%} &           80.19\% & \textbf{0.25m} & \textbf{2.77$^\circ$} & \textbf{36.98\%} &          86.69\% &           93.20\% & \textbf{0.11m} & \textbf{1.58$^\circ$} &           6.09ms \\
 FGR                   &           7.51\% &          52.28\% &           72.31\% &          0.51m &          8.71$^\circ$ &          10.06\% &          70.12\% &           86.69\% &          0.26m &          7.76$^\circ$ &          14.56ms \\
 FGR+p2p               &          22.90\% &          64.95\% &           77.69\% &          0.49m &          7.57$^\circ$ &          35.80\% &          87.28\% &           94.08\% &          0.25m &          5.70$^\circ$ &          20.76ms \\
 Ours                  &           6.41\% &          50.96\% &  \textbf{82.77\%} &          0.31m &          8.53$^\circ$ &           9.76\% &          65.38\% &           94.08\% &          0.16m &          3.47$^\circ$ &  \textbf{1.22ms} \\
 Ours+p2p              &          21.65\% &          64.95\% &           81.74\% &          0.31m &          8.66$^\circ$ &          35.21\% & \textbf{87.87\%} &  \textbf{94.38\%} &          0.14m &          3.57$^\circ$ &           7.00ms \\
 \hline
 Ours from scratch     &           5.45\% &          53.98\% &           73.05\% &          0.36m &          7.55$^\circ$ &           7.10\% &          73.37\% &           87.87\% &          0.17m &          3.39$^\circ$ &  \textbf{1.22ms} \\
 Ours from scratch+p2p &          20.03\% &          62.81\% &           72.97\% &          0.36m &          8.07$^\circ$ &          32.25\% &          82.54\% &           87.87\% &          0.16m &          3.09$^\circ$ &          12.14ms \\
\hline
\end{tabular}

  \vspace{-6pt}
  \caption{Registration results on \texttt{KITTITrackletsCars}, containing consecutive LiDAR scans of cars, extracted from KITTI tracking~\cite{Geiger12CVPR}.}
  \label{tab:KITTITrackletsCars}
\end{table*}
\begin{table*}[t]
  \centering\footnotesize\begin{tabular}{l|rrrrr|rrrrr|r}
  \multicolumn{1}{c}{} & \multicolumn{5}{c}{distance $<$ 80m} & \multicolumn{5}{c}{distance $<$ 20m} \\
  \hline
                &    2cm,1$^\circ$ &   10cm,5$^\circ$ &   20cm,10$^\circ$ &         RMSE $t$ &                 RMSE $R$                & 2cm,1$^\circ$    & 10cm,5$^\circ$   & 20cm,10$^\circ$   & RMSE $t$ &                 RMSE $R$                & time/transform       \\
\hline
 Global ICP     &          15.47\% &          68.25\% &           81.44\% &          0.53m &          5.66$^\circ$ &          15.44\% &          76.17\% &           90.72\% &          0.29m &          5.14$^\circ$ &        1593.29ms \\
 Global ICP+p2p &          23.05\% &          72.02\% &           83.23\% &          0.52m &          5.34$^\circ$ &          24.38\% &          82.89\% &           94.52\% &          0.28m &          4.03$^\circ$ &        1597.91ms \\
\hline
 ICP p2p        & \textbf{22.43\%} & \textbf{70.28\%} &  \textbf{84.78\%} & \textbf{0.20m} & \textbf{4.06$^\circ$} & \textbf{25.95\%} & \textbf{83.11\%} &  \textbf{95.30\%} & \textbf{0.07m} & \textbf{3.41$^\circ$} &           5.45ms \\
 FGR            &           9.82\% &          53.68\% &           72.68\% &          0.40m &         11.11$^\circ$ &          11.07\% &          60.63\% &           79.64\% &          0.14m &         11.46$^\circ$ &          15.08ms \\
 FGR+p2p        &          19.73\% &          66.97\% &           81.43\% &          0.38m &          7.73$^\circ$ &          23.15\% &          80.65\% &           93.29\% &          0.10m &          5.62$^\circ$ &          20.45ms \\
 Ours           &           3.77\% &          45.48\% &           66.51\% &          0.29m &         22.02$^\circ$ &           4.70\% &          42.51\% &           59.17\% &          0.22m &         27.73$^\circ$ &  \textbf{1.22ms} \\
 Ours+p2p       &          16.24\% &          58.29\% &           71.19\% &          0.29m &         18.98$^\circ$ &          17.56\% &          60.74\% &           70.47\% &          0.20m &         22.22$^\circ$ &           7.30ms \\
\hline
\end{tabular}

  \vspace{-6pt}
  \caption{Registration results on \texttt{KITTITrackletsCarsPersons}.}
  \label{tab:KITTITrackletsCarsPersons}
\end{table*}

\section{Experimental Evaluation}

To study object registration of partially observed objects captured by a LiDAR, we first generate a synthetic dataset by sampling points from CAD models.
In this controlled environment we can evaluate how well we can align point clouds, sampled at different distances and viewpoints. %
To evaluate our method on the real data, we extract LiDAR segments using the KITTI tracking dataset~\cite{Geiger12CVPR}.
We compare our methods against the state of the art geometric registration methods Global ICP~\cite{Rusu09ICRA} and FGR~\cite{zhou2016fast}, as well as centroid-initialized local ICP~\cite{Besl92PAMI} and ADH~\cite{Held14RSS}.

\subsection{Datasets}\label{sec:datasets}

\PAR{Synthetic Scenes.}
To obtain an extensive and diverse dataset of objects in arbitrary poses, we have created the synthetic dataset \texttt{SynthCars} specifically to study 3D registration of LiDAR scans.
To capture the characteristics of objects observed by a LiDAR sensor in street scenes, we simulate intersections of the rays characteristic to the Velodyne 64E laser scanner\footnote{\footnotesize{The same LiDAR sensor was used to record KITTI.}} using CAD models from the ModelNet40~\cite{Wu15CVPR} dataset.
We have hand-picked 100 car meshes, which are transformed using the orientation estimate provided by~\cite{Sedaghat17BMVC, Sedaghat15ICCV}.
Each mesh is additionally normalized by moving it to its centroid, and by uniformly scaling it to a maximum axis-aligned extent of 1.
Each of the $10\,000$ synthetic scenes is created by placing one mesh at a random pose in 2--80m distance from the origin (on the ground plane), with a random heading around the $z$-axis.
The normalized mesh is scaled by a random factor between 2.5--4.5m. 
A second mesh is placed at up to 1m distance to the first, with up to 90\deg relative rotation around the $z$-axis.
8000 scenes are created for the training set, with each mesh randomly chosen from the first 50 meshes; 1000 are created for validation and test sets, respectively, with the other 50 meshes.
To simulate LiDAR noise, we add Gaussian noise sampled from $\mathcal N_3(\vec0, \text{diag}\left(\sigma\right))$ to each point, with $\sigma=\max\{0.005, 0.05 \cdot d / 80\}$ at distance $d$, clipped at~$\left[-0.05, 0.05\right]$.

To demonstrate that our method can align a variety of object classes (even object classes not seen during training), we have created \texttt{SynthCarsPersons}. %
Here we pick a person mesh with 20\% probability (with a scale from 1.6--2m) and a car mesh otherwise.
We use 40 hand-picked meshes for the training set and 40 others for the validation/test sets.
We create \texttt{Synth20} by picking meshes randomly from the first 20 classes of ModelNet40, with a scale of 1--5m.
For each class we use the 20 first meshes for training and the remaining 20 for validation.
Similarly we create \texttt{Synth20others} from the remaining 20 classes to study the transfer to unseen classes.
We split all synthetic datasets into 8000/1000/1000 scenes for training, validation and testing.

\PAR{KITTI.}
We use KITTI track labels in form of 3D bounding boxes relative to the ego-vehicle in order to obtain 3D segments of objects, recorded in street scenes. They provide us with both training labels and point cloud segmentation (we extract all points within labeled 3D boxes).
To compensate for the slight differences in height of the objects scans, we transform both point sets to a common ground plane.

For \texttt{KITTITrackletsCars} we extract point clouds of the car and van category from successive frames of the 21 KITTI tracking training sequences (this excludes objects that re-appear after more than one frame). 
The sequences are divided into training and validation sequences, as in \cite{Voigtlaender19CVPR}. Sequences $\{2, 6, 7, 8, 10, 13, 14, 16, 18\}$ are further split into set $\{13, 14, 16, 18\}$ for the validation and $\{2, 6, 7, 8, 10\}$ for the test set, resulting in 20518/7432/1358 non-empty scenes.
Similarly we create \texttt{KITTITrackletsCarsPersons} using car, van and person classes. It contains 28463/10003/2069 scenes.
The KITTI dataset is captured at 10 Hz, therefore frame-to-frame pose and viewpoint changes are minor.
We therefore create the more challenging variants \texttt{KITTITracklets\-Cars\-Hard} and \texttt{KITTITracklets\-Cars\-Persons\-Hard} by taking object scans with large temporal gaps -- 10 or more frames difference and a minimum of 45\deg rotation around the $z$-axis.

\subsection{Baselines}

We compare our method against a set of local and global point cloud registration methods. We have simplified the registration task to a 2D translation and a $z$-axis rotation. To ensure a fair comparison we have constrained all baseline methods to only predict transformations of this limited search space.

\PAR{Local ICP.}\label{sec:baselinep2p}
For a simple, yet common baseline, we use the point-to-point ICP algorithm~\cite{Besl92PAMI} implemented in Open3D~\cite{Zhou18arxiv}. We have modified the minimization step by projecting all corresponding points to the ground plane, so that only rotations around the $z$-axis are computed. We initialize the transform with a translation aligning the point cloud centroids, and use a search radius of 0.1m for finding point correspondences. This configuration is also used for refining the initialization provided by the global registration methods (denoted in all tables by \emph{+p2p}).

\PAR{Global ICP.}
For the baseline of Global ICP~\cite{Rusu09ICRA}, we again use an implementation of Open3D~\cite{Zhou18arxiv} with its recommended configuration. Due to its computation time, this method is not applicable to real-time tracking. It is nonetheless interesting to analyze its registration performance. We therefore report all numbers, but do not compare them in detail to the other methods.

\PAR{Fast Global Registration.}
Fast Global Registration (FGR)~\cite{zhou2016fast} serves as a strong baseline, providing good initializations for a local ICP refinement, although still being an order of magnitude slower than our method. It also uses FPFH features, computed as in Global ICP~\cite{Rusu09ICRA}.

\subsection{Evaluation Protocol}

\PAR{Synthetic Data and KITTI Tracklets.}
As we restrict our task to proposing a $xy$-plane translation and a $z$-axis rotation, we evaluate these predictions and their errors separately. We compute the root mean squared error (RMSE) for the respective deviations from the ground truth.
Additionally, we evaluate both jointly to investigate how many of the predicted transforms are correct under the regime of different translational and rotational thresholds.
We therefore report the number of correct predictions within the translation/angle bins 2cm,1\deg; 10cm,5\deg and 20cm,10\deg. %

For the datasets restricted to the car category we noticed that the absolute angle deviation often converged to about 90\deg. That is because cars, especially sparsely sampled cases due to a large distance, appear symmetric along their heading axis. This is not only a problem during evaluation, where cars with indistinguishable heading can lead to large angular errors (180\deg).
It is also a problem for the loss computation during training. Therefore on all datasets with cars and persons we train our networks with the loss computed with the minimum angle to the orientation axis. During evaluation we also only take the smallest angle to the orientation axis as angle deviation. Only on \texttt{Synth20} and \texttt{Synth20others} we train and evaluate for aligning the actual object headings.

When objects are recorded using a high frame rate, no object is expected to turn more than 90\deg from frame to frame. Therefore the predicted object heading axes can be trivially converted to object headings by flipping all orientation predictions with more than 90\deg difference. For the more challenging dataset \texttt{KITTITrackletsCarsHard}, where this assumptions does not hold, a classification output could be added to predict whether to flip the prediction heading. To overcome the symmetry of sparse car scans, the associated image patch could be used to classify whether the object points into one direction or the other.

To separately evaluate the performance of all methods in close proximity, where a precise alignment is especially important, we additionally report all numbers for the subset of scenes that are within a 20m range (specifically, where the ground truth center of the first point cloud is at most 20m away from the scanner position).

\subsection{Discussion}

\PAR{Synthetic Scenes.}
Our method produces high-quality initial alignments in just over a millisecond per transform. This is (on all datasets) at least 10 times faster than the fast global registration method FGR.
For \texttt{SynthCars} and \texttt{SynthCarsPersons} our method obtains the lowest RMSE compared to the geometric baselines, for far a away and close by objects (see Tables~\ref{tab:SynthCars} and~\ref{tab:SynthCarsPersons}).
Moreover, we achieve the best results in all but the hardest translation/angle bins for \texttt{SynthCars}. A refinement with local ICP is able to produce very precise alignments for good initializations (manifesting as a high percentage of alignments within 2cm/1\deg), while the less precise initializations within 20cm/10\deg can diverge to local minima and have a negative impact on the performance. %

Experiments on \texttt{Synth20} and \texttt{Synth20others} show that our method is able to generalize to a large variety of object classes. As can be seen in Tables~\ref{tab:Synth20} and~\ref{tab:Synth20others}, our models (with refinement) perform about as good or better than the baselines within the 10cm/5\deg and 20cm/10\deg thresholds. Here we evaluate the angle deviation to the actual heading, not just the heading axis. For this experiment we have trained models with a loss function that penalizes the angle to the actual heading angle.
The root mean squared angle deviation of our method is about 90\deg, indicating that our method has problems predicting the orientation of these objects. Several classes in ModelNet dataset are symmetric (\eg bathtub, bed, desk) and the additional sub-sampling required for our approach increases the orientation ambiguity.

\PAR{KITTI Tracklets.}
For all datasets generated from KITTI recordings we pre-train our models on the respective synthetic datasets. %
As can be seen in Tables~\ref{tab:KITTITrackletsCars} and~\ref{tab:KITTITrackletsCarsPersons}, the results of all methods indicate that for \texttt{KITTITrackletsCars} and \texttt{KITTITrackletsCarsPersons}, a local geometric method is fast and very accurate.
We attribute this to the fact that frame-to-frame displacements (recorded under high frame rate) are relatively small, especially rotations.
The significant overlap of successive point clouds forms perfect conditions for the local ICP method, which eliminates the need for global registration methods.
Nonetheless, on \texttt{KITTITrackletsCars}, our method shows similar or better performance compared to FGR. With the local refinement, we achieve the best overall results for some of the less challenging translation/angle bins.
In Table~\ref{tab:KITTITrackletsCars} we additionally show results without pre-training on \texttt{SynthCars}, which indicate that pre-training boosts the performance on \texttt{KITTITrackletsCars}, most notably within the 20cm/10\deg thresholds.
For \texttt{KITTITrackletsCarsPersons}, the best performing method is FGR. The higher angular RMSE error (compared to \texttt{KITTITrackletsCars}) indicates that our method occasionally fails to identify a person's heading correctly.

More challenging settings for \texttt{KITTITracklets\-CarsHard} and \texttt{KITTITracklets\-Cars\-Persons\-Hard} with larger translation and rotation gaps lead to vastly different angles of observation and point cloud densities. In this case all geometric methods struggle to correctly align the point clouds. In contrast, our learning-based method can learn to be robust against varying densities and larger gaps. %
A refinement of our predictions with local ICP leads to lower scores, implying that these scenarios are so challenging that ICP cannot correctly align point clouds even when a coarse initialization is provided. %

We conclude that when relative motion and point cloud density does not change significantly even local registration methods perform comparatively well to our approach.
Our method performs significantly better in harder scenarios with larger temporal gaps.
We note that even though harder cases appear infrequently in a real-world setting, it is important that tracking methods are robust to such cases as failures can lead to catastrophic consequences. %

\begin{table}[t]
  \centering\setlength{\tabcolsep}{3pt}\footnotesize\begin{tabular}{l|rrrrr}
  \multicolumn{1}{c}{} & \multicolumn{5}{c}{distance $<$ 80m} \\
\hline
                              &   2cm,1$^\circ$ &   10cm,5$^\circ$ &   20cm,10$^\circ$ &         RMSE $t$ &                 RMSE $R$ \\
\hline
 Global ICP     &          0.00\% &           0.00\% &            0.00\% &         38.34m &          70.78$^\circ$ \\
 Global ICP+p2p &          0.00\% &           0.00\% &            0.00\% &         38.34m &          70.76$^\circ$ \\
\hline
 ICP p2p        &          0.00\% &           0.00\% &            0.00\% &          0.51m &          68.66$^\circ$ \\
 FGR            &          0.00\% &           0.00\% &            0.00\% & \textbf{0.48m} &          70.82$^\circ$ \\
 FGR+p2p        &          0.00\% &           0.00\% &            0.00\% &          0.51m &          68.96$^\circ$ \\
 Ours           &          0.00\% &  \textbf{1.80\%} &  \textbf{14.48\%} & \textbf{0.48m} & \textbf{31.08$^\circ$} \\
 Ours+p2p       & \textbf{0.05\%} &           1.14\% &            9.44\% &          0.52m &          33.11$^\circ$ \\
\hline
\end{tabular}

  \vspace{-6pt}
  \caption{Registration results on \texttt{KITTITracklets\-CarsHard}, with cars captured with more than 10 frames difference and at least 45\deg angle difference.}
\end{table}
\begin{table}[t]
  \centering\setlength{\tabcolsep}{3pt}\footnotesize\begin{tabular}{l|rrrrr}
  \multicolumn{1}{c}{} & \multicolumn{5}{c}{distance $<$ 80m} \\
\hline
                &   2cm,1$^\circ$ &   10cm,5$^\circ$ &   20cm,10$^\circ$ &         RMSE $t$ &                 RMSE $R$ \\
\hline
 Global ICP     &          0.00\% &           0.00\% &            0.00\% &         35.16m &          69.31$^\circ$ \\
 Global ICP+p2p &          0.00\% &           0.00\% &            0.05\% &         35.16m &          69.27$^\circ$ \\
\hline
 ICP p2p        &          0.00\% &           0.00\% &            0.05\% &          0.48m &          67.91$^\circ$ \\
 FGR            &          0.00\% &           0.00\% &            0.00\% & \textbf{0.46m} &          69.61$^\circ$ \\
 FGR+p2p        &          0.00\% &           0.00\% &            0.05\% &          0.49m &          68.23$^\circ$ \\
 Ours           &          0.00\% &  \textbf{1.78\%} &  \textbf{11.16\%} & \textbf{0.46m} & \textbf{38.63$^\circ$} \\
 Ours+p2p       &          0.00\% &           1.45\% &            8.77\% &          0.49m &          40.56$^\circ$ \\
\hline
\end{tabular}

  \vspace{-6pt}
  \caption{Registration results on \texttt{KITTITracklets\-CarsPersonsHard.}}
\end{table}

\begin{table}[t]
  \centering\footnotesize\begin{tabular}{l|ccc|r}
\multicolumn{1}{c}{}&  \multicolumn{1}{c}{d. $<$ 80m}  &\multicolumn{1}{c}{d. $<$ 20m}&   \multicolumn{1}{c}{d. $<$ 5m} &     \\
\hline
&    RMSE $v$ & RMSE $v$ &   RMSE $v$ &     time/transf. \\
\hline
 Centr. Kalman Filter      &          2.568m/s &          1.251m/s &          1.604m/s & \textbf{0.004ms} \\
 ADH 2D~\cite{Held14RSS}            &          2.691m/s &          1.139m/s &          1.023m/s &          0.253ms \\
 ADH 2D (parallel) &          2.691m/s &          1.139m/s &          1.023m/s &          0.066ms \\
 ADH 3D~\cite{Held14RSS}            &          2.682m/s &          1.132m/s &          1.002m/s &          0.418ms \\
 Ours               & \textbf{1.834m/s} & \textbf{0.851m/s} & \textbf{0.732m/s} &          1.220ms \\
\hline
\end{tabular}

  \vspace{-6pt}
  \caption{Comparison of our method to ADH~\cite{Held14RSS} and centroid-based Kalman filter.}
  \label{tab:eval_adh}
\end{table}

\PAR{Comparison to ADH.}
To compare to ADH~\cite{Held14RSS} and their baseline (centroid-based Kalman filter) we have extended their evaluation script to our \texttt{KITTITrackletsCars} dataset.
For this evaluation, only velocity needs to be estimated.
We simply adapt our inference to output frame-to-frame velocities from the predicted translations, assuming known time difference between two scans.
We average each translation with those of the two adjacent frames to smoothen out small translation errors.
Note that all variants of~\cite{Held14RSS} use a motion model, whereas our method does not condition its estimate on the past evidence.
Table~\ref{tab:eval_adh} summarizes our results.
Our method outperforms variants and the baseline of~\cite{Held14RSS} by a considerable margin in terms of RMSE (0.732m/s vs. 1.002m/s in the 5m range), especially for objects observed in farther ranges (1.834m/s vs. 2.682m/s). However, our method is slower.
We notice that the simple Kalman filter baseline is more accurate compared to ADH for objects with up to 80m distance, however in the near-range ADH is significantly more accurate.

\subsection{Timings}
All of our timings have been conducted on a machine with an Intel i7-3770 CPU, an Nvidia GTX 1080 Ti and 32 gigabytes of RAM. For all timings, only the computation time was measured, excluding time for data loading. All ICP baselines run multi-threaded on the CPU. The reported times are averaged over the whole test set. Our method can run most operations on the GPU, benefiting from massive parallelization. The time per transform is computed by dividing the mean batch processing time by the batch size. %
Our method is most efficient for an inference batch size of 32, with about 1.22ms per alignment (this is the reported time in all our tables). It is also a realistic number of objects to track simultaneously. Batch sizes 8, 16 and 64 are not much slower with 1.93ms, 1.45ms and 1.34ms.

\section{Conclusion}

We have presented an efficient, data-driven point cloud registration algorithm for real-time object tracking applications. Our siamese network predicts a coarse alignment by transforming both point clouds to an amodal canonical pose and concatenates point cloud embeddings to predict a final alignment. We experimentally demonstrated that our method is computationally efficent and robust to large rotations and significantly different point densities. %
In future work, we plan to investigate whether more sophisticated point cloud processing architectures such as~\cite{Qi17NIPS, Wang19TOG} could improve our performance and to perform sequence level learning using recurrent neural networks. %

\vspace{4pt}
\footnotesize {\PAR{Acknowledgments:} This project was funded, in parts, by ERC Consolidator
Grant DeeVise (ERC-2017-COG-773161). We would like to thank Alexander Hermans, Istv\'an S\'ar\'andi and Paul Voigtlaender for productive discussions.}

{\small
\bibliographystyle{ieee}
\bibliography{abbrev_short,pclalign}
}

\end{document}